\documentclass[runningheads]{llncs}
\usepackage[T1]{fontenc}
\usepackage{graphicx}
\usepackage{booktabs}
\usepackage[dvipsnames]{xcolor}
\usepackage[most]{tcolorbox}
\usepackage{amsmath}
\usepackage{amsfonts,amssymb}
\usepackage{authblk}
\usepackage[misc]{ifsym} 
\usepackage{caption}
\usepackage{subcaption}
\usepackage{bbding}
\usepackage[labelfont=bf]{caption}
\usepackage{color}
\usepackage{colortbl}
\usepackage{hyperref}
\usepackage{pifont}
\usepackage[nameinlink,capitalize]{cleveref}

\newcommand\blfootnote[1]{%
  \begingroup
  \renewcommand\thefootnote{}\footnotetext{#1}%
  \addtocounter{footnote}{-1}%
  \endgroup
}
\hypersetup{
colorlinks=true,
linkcolor=blue,
anchorcolor=black,
citecolor=blue
}

\definecolor{mygray}{gray}{.9}
\begin{document}
\title{Incorporating Clinical Guidelines through Adapting Multi-modal Large Language Model for Prostate Cancer PI-RADS Scoring}
\titlerunning{Incorporating PICG}
%
\author{Tiantian Zhang\inst{1\dag} \and 
Manxi Lin\inst{2\dag} \and 
Hongda Guo \inst{3} \and
Xiaofan Zhang \inst{4,5} \and
Ka Fung Peter Chiu \inst{3} \and
Aasa Feragen \inst{2} \and 
Qi Dou \inst{1(\textrm{\Letter})}
}
\authorrunning{T. Zhang, M. Lin, et al.}

\institute{Department of Computer Science and Engineering, The Chinese University of Hong Kong, Hong Kong, China \\ \and 
Department of Applied Mathematics and Computer Science, Technical University of Denmark, Kongens Lyngby, Denmark \and 
Department of Surgery, The Chinese University of Hong Kong, Hong Kong, China \and Shanghai Jiao Tong University, Shanghai, China \and 
Shanghai Artificial Intelligence Laboratory, Shanghai, China}

\maketitle              
\begin{abstract}
The Prostate Imaging Reporting and Data System (PI-RADS) is pivotal in the diagnosis of clinically significant prostate cancer through MRI imaging. Current deep learning-based PI-RADS scoring methods often lack the incorporation of common PI-RADS clinical guideline~(PICG) utilized by radiologists, potentially compromising scoring accuracy. 
This paper introduces a novel approach that adapts a multi-modal large language model (MLLM) to incorporate PICG into PI-RADS scoring model without additional annotations and network parameters.
We present a designed two-stage fine-tuning process aiming at adapting a MLLM originally trained on natural images to the MRI images while effectively integrating the PICG. Specifically, in the first stage, we develop a domain adapter layer tailored for processing 3D MRI inputs and instruct the MLLM to differentiate MRI sequences. In the second stage, we translate PICG for guiding instructions from the model to generate PICG-guided image features. Through such a feature distillation step, we align the scoring network's features with the PICG-guided image features, which enables the model to effectively incorporate the PICG information. We develop our model on a public dataset and evaluate it on an in-house dataset. 
Experimental results demonstrate that our approach effectively improves the performance of current scoring networks. Code is available at: \url{https://github.com/med-air/PICG2scoring}
\end{abstract}
\section{Introduction}
\blfootnote{
${}^\dag$ Equal contribution (emails: tiantianzhang@cuhk.edu.hk, manli@dtu.dk)\\
}
The adoption of MRI has helped on better prostate cancer diagnosis, especially for clinically significant stages. The Prostate Imaging Reporting and Data System (PI-RADS) plays a crucial role in standardizing the MRI report when assigning the score to a suspicious lesion. In practice, the PI-RADS evaluation follows the well-acknowledged PI-RADS Clinical Guideline (\textbf{PICG})~\cite{bickle_2023}, with which clinicians grade the MRI images according to the morphological characteristics of lesions. 
Specifically, the PICG involves the assessment of three common MRI sequences: T2-weighted imaging~(T2W), apparent diffusion coefficient~(ADC), and diffusion-weighted imaging~(DWI). According to the PICG, each identified lesion is assigned a PI-RADS score ranging from 1 to 5, a higher score indicates higher risk of prostate cancer~\cite{ahmed2017diagnostic,park2016prostate,purysko2021pi}.

Recently, AI-assisted models have emerged as a prominent tool for automatic PI-RADS scoring. Current methods~\cite{gravina2022machine,gu2023deep,kang2024deep,schelb2019classification} formulate the task as a classification problem, where the model is optimized to discrete labels. Although these labels are annotated based on PICG, the models typically only learn the scores rather than understanding the actual guidelines. 
This can lead to several issues, including the exposure to subjective biases during the labeling process, as different annotators may interpret the guidelines unconsciously different~\cite{kafkalias2022bias}. As a result, the model may not fully capture the nuanced information embedded in the guidelines, potentially limiting its trustworthiness in real-world use.

Informing the networks with annotation guidelines is promising to alleviate the challenge. A straight-forward solution is to design rule-based models that translate clinical guidelines into part of the classification network~\cite{koh2020concept,lin2022saw,yang2023language}.
However, these methods often require network architecture modifications or the inclusion of additional labels~\cite{lin2022saw}. To address these limitations, we introduce the concept of feature distillation to our model design. Specifically, we build a model on both images and PICG, namely the \textit{guideline network}, which learns to regularize the image feature space with PICG. Given an arbitrary \textit{scoring network}, we transfer the learned PICG information by aligning the feature distribution of the scoring network with the PICG-regularized features from the guideline network. Our approach allows seamless integration of PICG information into various scoring networks without extensive architectural changes.

We consider our guideline network as a model handling both image and text inputs. Multi-modal Large Language Models (MLLMs) offer a state-of-the-art example of text and image integration~\cite{alayrac2022flamingo,dai2023instructblip,gao2023llama,wu2023multimodal}. In medical imaging, MLLMs have been instructed to perform tasks such as medical visual question answering~\cite{li2024llava,zhang2023pmc}. 
Different from the former works, the text input in our guideline network, i.e., PICG, serves as an instruction for regularizing the input image feature space.

In this paper, we propose an approach for incorporating PICG into the scoring network through the guideline network without modifying the network structure or requiring additional training data. We design a two-stage fine-tuning strategy to adapt the guideline network, an MLLM, to inject the PICG information into the 3D MRI image features. Leveraging feature distillation, we align the features from the scoring and guideline networks, hereby incorporating PICG to enhance the model performance. 
In the first stage, we design a domain adapter layer to facilitate the model's support for 3D MRI input and then adjust the instructions of the guideline network to aid in distinguishing MRI sequences. These instructions offer detailed descriptions for different sequences. In the second stage, we transform PICG into instructions for the guideline network. We guide the model to generate PICG-guided image features 
by replacing each PI-RADS label with its corresponding PICG sections. 
Training is conducted on a public dataset~\cite{natarajan2020prostate}, and we perform testing on a private dataset. We select three state-of-the-art scoring methods as our scoring network: Yu et al.~\cite{yu2023pi}, Sanford et al.~\cite{sanford2020deep}, and Kang et al.~\cite{kang2024deep}. Experimental results indicate that incorporating PICG into these methods results in accuracy improvements of 4.8\%, 5.3\%, and 6.4\% over their original performance, respectively.

\section{Method}
\begin{figure}[t]
    \centering
    \includegraphics[width=0.95\linewidth]{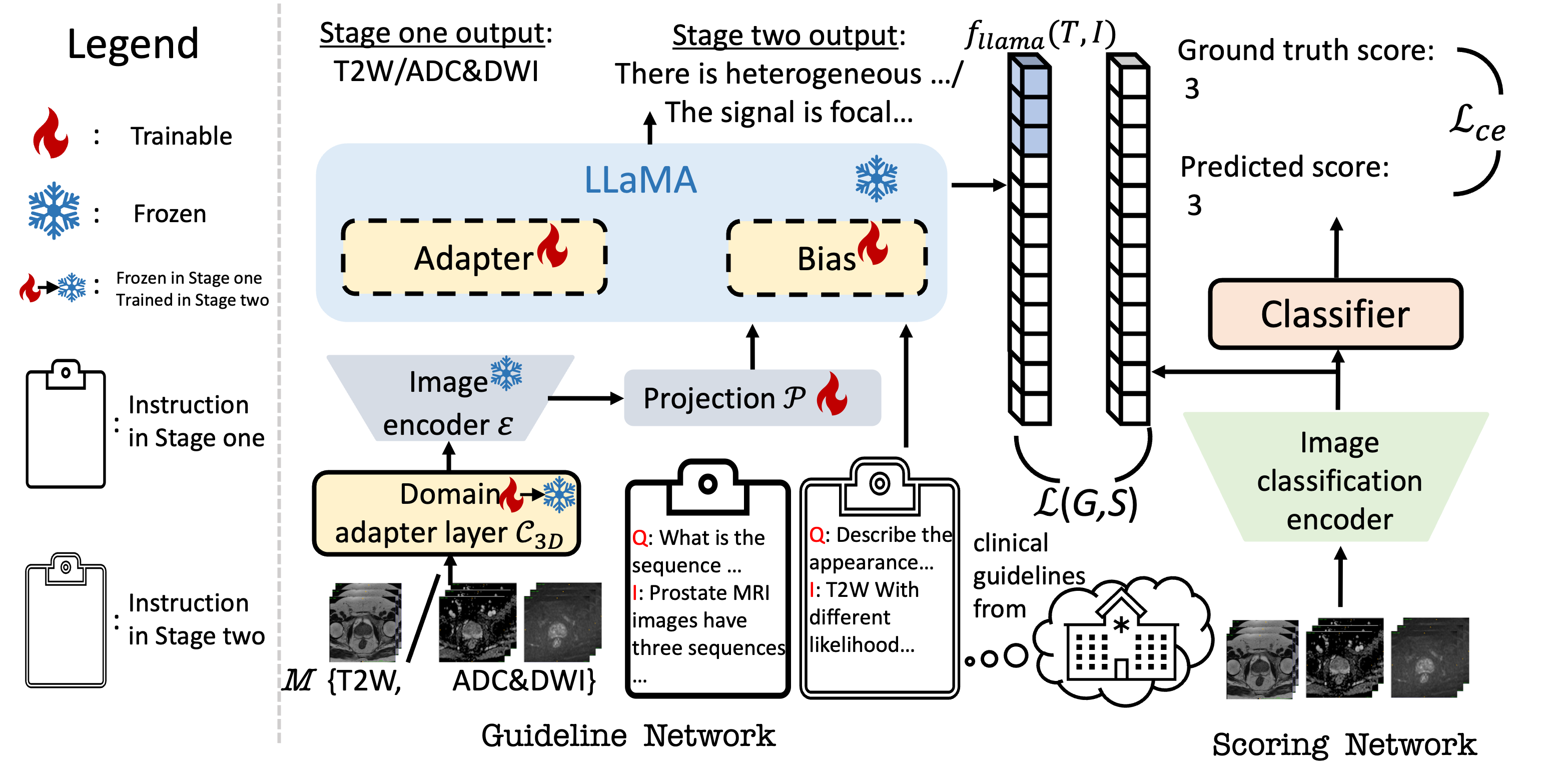}
    \caption{The overview of our proposed method which consists of two-stage instruction tuning and feature distillation. In stage one, we design and train the domain adapter layer and use instruction tuning to distinguish ``T2W'' and ``ADC\&DWI'' sequences. In stage two, we freeze the domain adapter layer and design another instruction to learn PICG. Domain adapter layer, image encoder, projection and LLaMA constitute the guideline network, while image classification encoder and classifier make up the scoring network.}
    \label{fig:network}
\end{figure}

Fig.~\ref{fig:network} shows the overview of our proposed PI-RADS scoring framework for incorporating PICG via guideline network, which uses a multi-modal large language model backbone. We first show the two-stage fine-tuning process. We establish the domain adapter and adjust the instructions to adapt guideline network to the MRI domain. Then, we elaborate on the instruction employed to direct guideline network in producing PICG-guided image features. At last, we show the feature distillation process from the guideline network to the scoring network.

\begin{figure}
    \centering
    \includegraphics[width=0.9\linewidth]{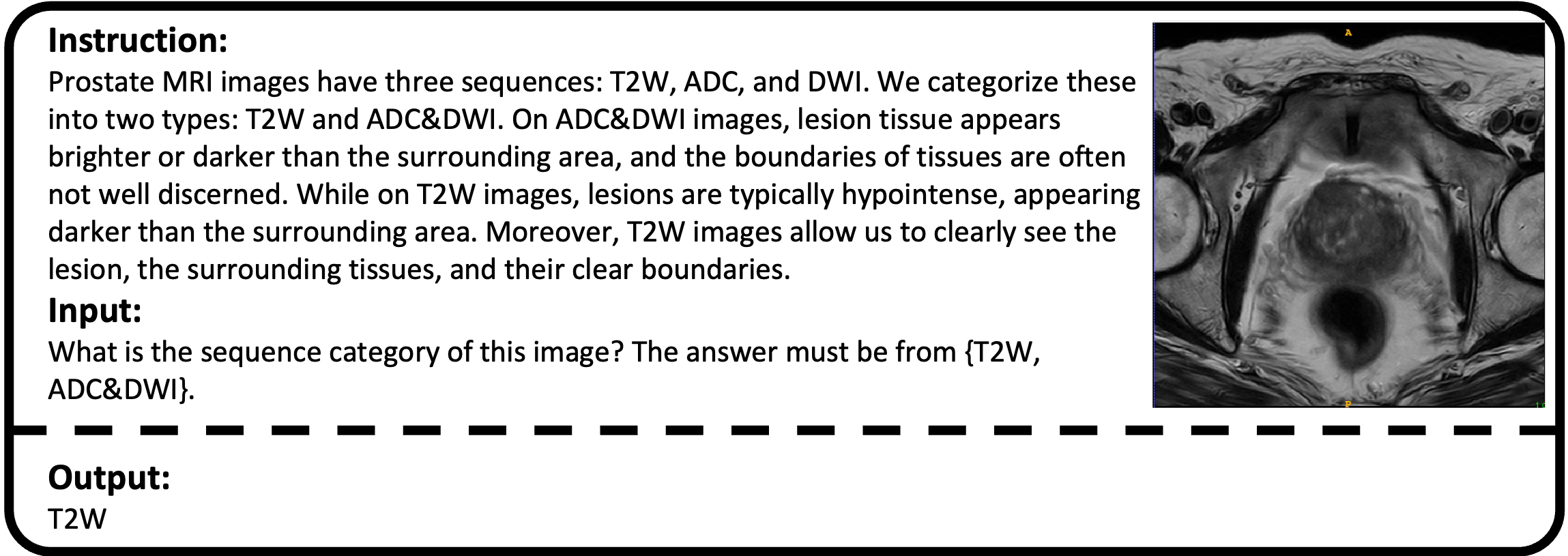}
    \caption{Example of instructions for adapting guideline network to Prostate MRI.}
    \label{instruction1}
\end{figure}

\subsection{Stage One: Adapting MLLM to Prostate MRI Domain}
Our guideline network is a multi-modal large language model (MLLM). First, we develop a domain adapter layer to enable the model to handle 3D MRI input. Following~\cite{zhang2023llama}, we utilize a pre-trained vision transformer~\cite{radford2021learning} as the image encoder. The features from the vision transformer are further processed by a multi-modal projection layer and then sent to the large language model LLaMA~\cite{gao2023llama}.

Our 3D volume includes an extra ``depth'' dimension compared to the 2D images on which the vision transformer is pre-trained. To address this, we adapt the patch embedding layer of the transformer to a 3D convolutional layer with a kernel size of 14, which we refer to as the domain adapter layer. We denote it as $\mathcal{C}_{3D}$. Following I3D~\cite{carreira2017quo}, we duplicate the pre-trained weights from the original embedding layer across the ``depth'' dimension to maintain the original patch division for each slice within the volume. As part of the pre-processing, the input volumes are reshaped to ensure the number of patches remains consistent with the original outputs of the vision transformer. Given the input 3D MRI image $M$, the visual prompts $I$ are encoded as:
\begin{equation}
        I = \mathcal{P}(\mathcal{E}(\mathcal{C}_{3D}(M))),\\
\end{equation}
where $\mathcal{E}$ is the rest blocks of the image encoder and $\mathcal{P}$ refers to the multi-modal projection layer. 
Following the approach in~\cite{gao2023llama}, the text input and instructions are processed and converted into text tokens. 
These text tokens are then combined with the visual prompt $I$ before being fed into LLaMA.

We use the instruction tuning to fine-tune the guideline network, enabling it to distinguish between ``T2W'' and ``ADC\&DWI'' images. T2W images provide anatomical detail, while ADC and DWI images reflect the diffusion properties of water molecules, offering different contrasts and diagnostic information~\cite{wang2014principles}. The instruction is shown in Fig.~\ref{instruction1} based on~\cite{gao2023llama}. Similar to~\cite{zhang2023llama}, we only fine-tune the patch embedding layer, the multi-modal projection layer, and the bias in the LLaMA model, while keeping all other trainable parameters frozen. This approach enables fine-tuning on a single GPU. Throughout the training process, the model is optimized using the cross-entropy loss function.

\begin{figure}[t]
    \centering
    \begin{subfigure}{0.9\linewidth}
    \includegraphics[width=\linewidth]{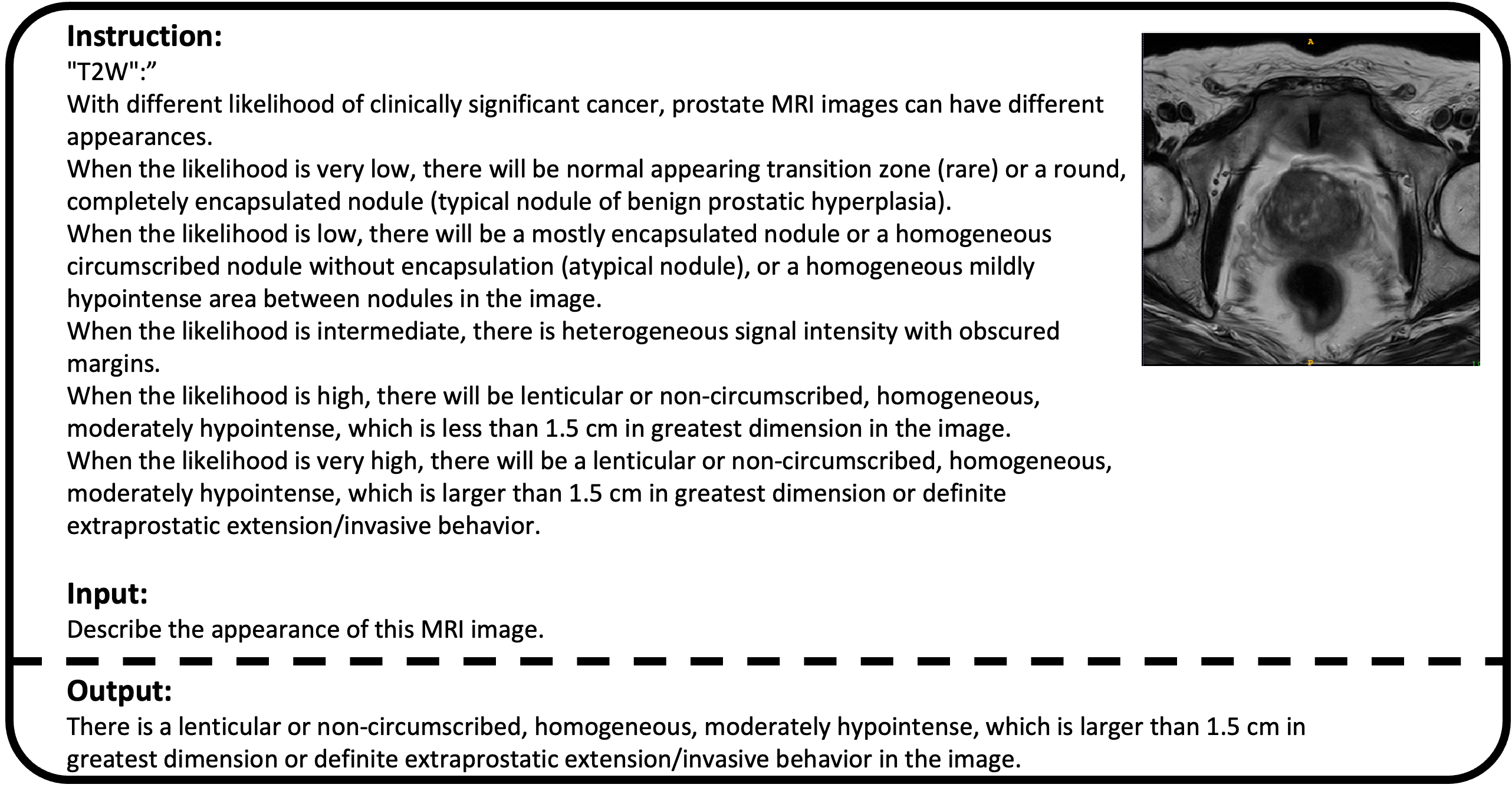}
    \end{subfigure}
    \begin{subfigure}{0.9\linewidth}
    \includegraphics[width=\linewidth]{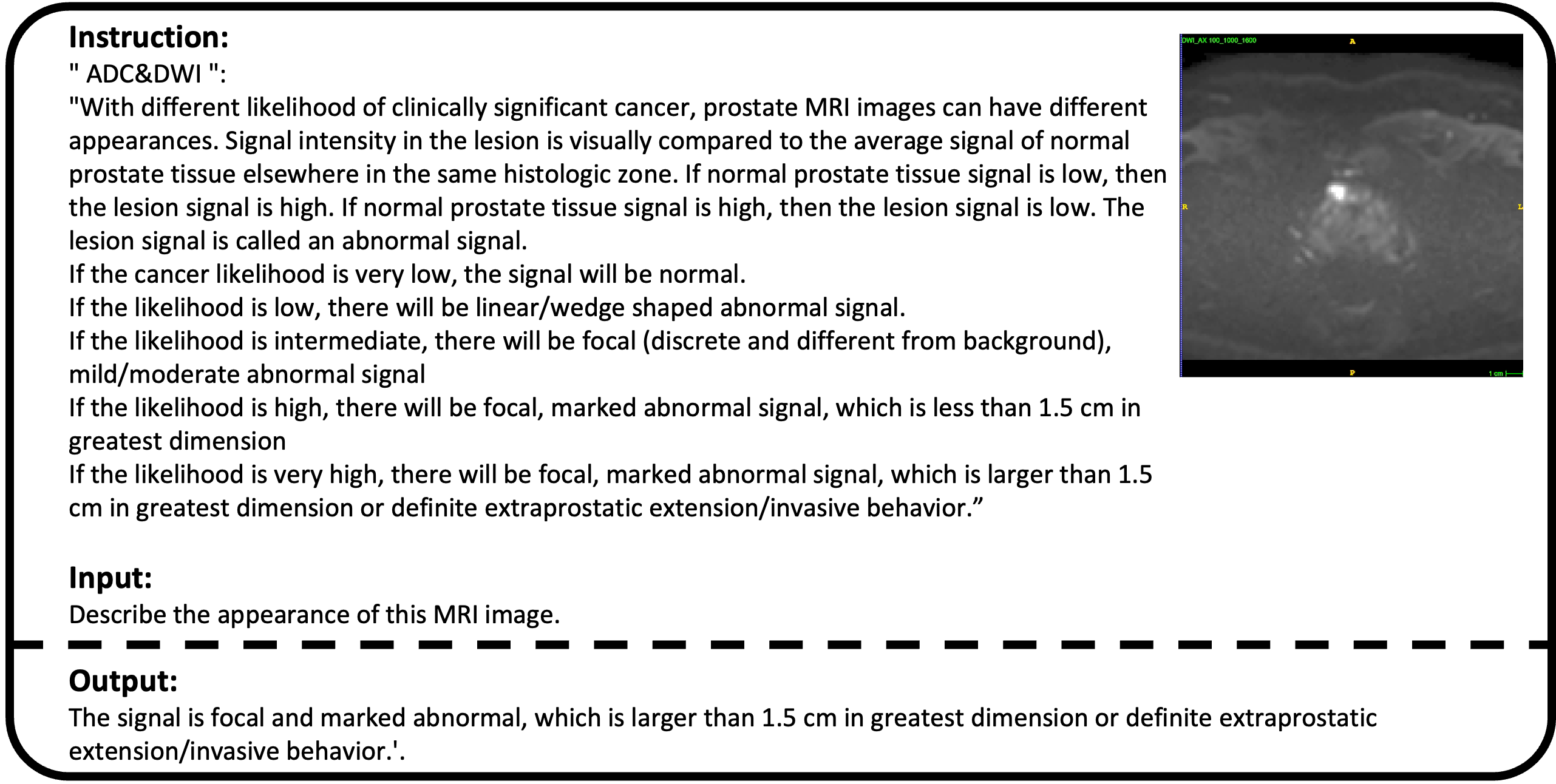}
    \end{subfigure}
    \caption{Examples of instructions used for generating PICG-guided image features.}
    \label{instruction2}
\end{figure}

\subsection{Stage Two: Generating PICG-guided Image Features}
We fine-tune the guideline network from the previous stage. During this stage, we freeze the domain adapter layer while allowing the multi-modal projection layer and the bias in the LLaMA model to remain trainable.
We instruct the guideline network to understand PICG. Our instruction for stage two is described in Fig.~\ref{instruction2}. We formulate the task as an image captioning task, where the model is asked to describe the lesion in the MRI image based on the PICG. This helps us to obtain an instruction-incorporated representation of the MRI image. Specifically, at the end of this stage, we utilize the fine-tuned model to generate PICG-guided features of each sample in the training set. We take the average of token features from the last transformer layer of the model to obtain PICG-guided image features. The feature, i.e., a representation of the MRI images that incorporates the PICG, will be leveraged to guide our PI-RADS scoring task. 

\subsection{PICG-guided Image Feature Distillation}
We incorporate the encoded PICG into the scoring network, which we denote as $S$, through a process of feature distillation. To achieve this, we first input the visual prompt $I$ and the text token $T$ into the LLaMA model, which is a part of the guideline network $G$. Then we extract the PICG-guided image feature, denoted as $f_{llama}(T,I)$, before the output stage of LLaMA. Subsequently, we obtain the features from the final layer of the scoring network $S$, denoted as $f_S(M)$. At last, the Kullback-Leibler (KL) divergence~\cite{liu2023functionconsistent} is used to distill the feature information as:

\begin{equation}
    \mathcal{L}(G, S) = \textbf{KL}(f_{llama}(T,I)\|\sigma(f_S(M))),\label{kl}
\end{equation}
where $\sigma $ is one fully connected layer only used in the feature distillation. We use $\sigma $ to align the feature size of the scoring network with that of the PICG-guided image feature. This layer will be discarded during inference.

\subsection{Learning Process}
In the first fine-tuning stage, we set the epoch number to 20, with a 2-epoch warmup. The weight decay is 0.02, and the learning rate is set to 0.02. In the second stage, we finetune the model from the first stage for 60 epochs. The warmup epoch number is 5. 
The overall objective function of the scoring network is as:~$  \ell = \mathcal{L}_{ce} + \alpha * \mathcal{L}(G,S)$
where $\mathcal{L}_{ce}$ is the cross-entropy loss. We use three representative PI-RADS scoring networks and incorporate the PICG into these networks. We leverage pre-training weights from the Kinetics-700~\cite{carreira2019short} and Moments in Times datasets~\cite{monfort2019moments}, set the batch size to 16, employ focal loss as the loss function with weights specified as [2, 2, 1, 1, 1], and $\gamma$ set to 2. Our optimization strategy involves using Adam with a learning rate of 5e-5 over 200 epochs. All models can be trained on a single NVIDIA A40 GPU card.

\section{Experiments}

\subsection{Datasets}

\subsubsection{Public Dataset.}
For model development, we employed the publicly available Prostate-MRI-US-Biopsy dataset~\cite{natarajan2020prostate}, which includes preoperative MRI images, lesion locations, and PI-RADS scores. This dataset contains 762 lesions from 615 subjects, with MRIs in T2W, ADC, and DWI sequences. Following~\cite{yu2023pi}, regions of interest were isolated based on location annotations, and $10\%$ of the data was reserved for validation (i.e., 683 cases for training, 79 for validation).

\subsubsection{Private Dataset.}
For external testing, we gathered a private dataset of T2W, ADC, and DWI sequence images from 206 patients, derived from authentic clinical procedures. This dataset was sourced from a prospectively maintained cohort of prostate biopsy patients at Prince of Wales Hospital in Hong Kong, including only those who underwent MRI-ultrasound fusion targeted biopsy for ground-truth labels. Although the biopsies were performed at the same medical center, the pre-biopsy MRI examinations were conducted at various medical centers across Hong Kong. Consequently, the cohort's images are heterogeneous, acquired using different scanners from three manufacturers (Philips, Siemens, GE) and two magnetic field strengths (1.5 T and 3 T). Despite this variability, all images were of sufficient quality for radiologists to assign PI-RADS scores to suspicious lesions and for urologists to perform MRI-ultrasound fusion targeted biopsies effectively. Two radiologists independently annotated 293 lesions with location and PI-RADS scores, and these lesions were identified and extracted for further analysis. 
In our experimental setting, the model was trained on public dataset and tested on private dataset, i.e., all the results are reported on this out-of-distribution private dataset.

\subsection{Comparison with State-of-the-Art Methods}
To demonstrate the effectiveness of our method, we compared it with the state-of-the-art PI-RADS scoring networks~\cite{kang2024deep,sanford2020deep,yu2023pi}, as presented in Table~\ref{tab:benchmark}. For a fair comparison, rule-based methods that require additional annotations were not included as the baselines. The evaluation metrics involve Classification Accuracy, Mean Squared Error (MSE), Mean Absolute Error (MAE), Precision, Recall, and F1 metrics. Precision, Recall, and F1 scores were averaged across all classes. The proposed knowledge distillation was performed to the encoded features of these scoring networks, as a plug-in. Following the application of our method, all three scoring networks and the VGG network show improvements across all metrics. Particularly, our method brings an improvement of 6.4\% in accuracy and 2.9\% in precision to~\cite{kang2024deep}. This highlights the effectiveness of integrating PICG into scoring networks for enhancing accuracy. Additionally, our method leads to a decrease in false positives. Furthermore, the incorporation of PICG resulted in reduced MSE and MAE values across all three networks, indicating that the generated scores are closer to the ground truth. It is important to note that our proposed approach is compatible with any scoring network, due to its flexibility.

\begin{table}[t]
\setlength{\tabcolsep}{3pt}
  \centering
  \caption{Performance of different methods on the private test set. Results are reported with the average and standard deviation over three independent runs. }
  \label{tab:benchmark}
  \begin{center}
  \resizebox{\textwidth}{!}{
  \begin{tabular}{l|c|c|c|c|c|c}
    \toprule
    Model &Accuracy \% $\uparrow$ & MSE $\downarrow$& MAE $\downarrow$ & Precision\%$\uparrow$& Recall\%$\uparrow$& F1\%$\uparrow$\\
    \hline
    VGG~\cite{simonyan2014very} & 31.6$\pm$1.4&1.38$\pm$0.2&0.92$\pm$0.5&17.4$\pm$1.5&22.6$\pm$0.8&17.4$\pm$1.8\\
    \rowcolor{mygray}VGG~\cite{simonyan2014very}+PICG (ours)&38.6$\pm$2.1(+7.0)&1.09$\pm$0.1&0.77$\pm$0.0&21.1$\pm$0.3&22.9$\pm$1.2&21.0$\pm$0.6\\
    \hline
    Kang et al.,~\cite{kang2024deep}& 30.0$\pm$4.0& 1.43$\pm$0.3&0.93$\pm$0.1 &13.2$\pm$2.6&20.1$\pm$2.5&14.5$\pm$2.0\\
    \rowcolor{mygray} Kang et al.,~\cite{kang2024deep}+PICG (ours)&36.4$\pm$1.0 (+6.4) & 1.25$\pm$0.0 & 0.83$\pm$0.0&16.1$\pm$3.6&20.9$\pm$1.8&15.9$\pm$2.3\\
    \hline
    Sanford et al.,~\cite{sanford2020deep}&30.4$\pm$0.4 & 1.61$\pm$0.2&0.97$\pm$0.1&17.7$\pm$1.3&22.1$\pm$1.7&16.1$\pm$1.1\\
    \rowcolor{mygray}Sanford et al.,~\cite{sanford2020deep}+PICG (ours)&35.7$\pm$0.9(+5.3)&1.38$\pm$0.2&0.87$\pm$0.1&18.4$\pm$1.2&22.0$\pm$0.5&17.4$\pm$1.5\\
    \hline
    Yu et al.,~\cite{yu2023pi}& 33.8$\pm$0.6&1.22$\pm$0.2&0.83$\pm$0.1&18.2$\pm$6.1&21.8$\pm$2.3&13.4$\pm$3.1\\
    \rowcolor{mygray}Yu et al.,~\cite{yu2023pi}+PICG (ours)& 38.6$\pm$0.4(+4.8)&1.17$\pm$0.1&0.79$\pm$0.0&20.4$\pm$0.6&23.8$\pm$1.4&20.6$\pm$0.8\\

    \bottomrule
  \end{tabular}
  }
  \end{center}
\end{table}

\subsection{Ablation Study}

\subsubsection{Determining Loss Function Weight $\alpha$.}

Large models have strong generalization capabilities, while small models excel at specific tasks, so we need to balance the proportion of loss weights.
We illustrate the experimental process of determining the optimal loss weight  $\alpha$ in Eq.~(\ref{kl}) as shown in Table~\ref{tab:my_label}. We assign three loss weights 0.2, 0.4, and 0.6, respectively. It is considerable that when the weight is set to 0.4, the model exhibits improved performance compared to when the weight is set to 0.2. However, increasing the weight to 0.6 results in a notable decrease in the model's accuracy. 

\subsubsection{The Effect of Two-stage Fine-tuning.}

We analyze the impact of fine-tuning MLLM on the scoring network's performance through a comparative experiment called ``baseline MLLM”. This experiment used MLLM in its initialized state without fine-tuning, maintaining the same input and instructions as the pre-trained model, and conducting feature distillation. As shown in Table~\ref{pre-train}, using features from the initialized MLLM (with ImageNet pre-trained weights) decreased accuracy compared to using fine-tuned features. The accuracy was also lower than the results reported by Sanford et al. \cite{sanford2020deep}. Similarly, following findings by Yu et al. \cite{yu2023pi} and Kang et al. \cite{kang2024deep}, using features without pre-training reduced the model's accuracy.
\begin{table}[h]
    \caption{Model performance with different $\alpha$. The best performance over three independent runs is reported.)}
    \label{tab:my_label}
\setlength{\tabcolsep}{3pt}
    \centering
      \begin{center}
      \resizebox{0.9\textwidth}{!}{
    \begin{tabular}{c|c|c|c|c|c|c|c|c}
        \toprule
        \multicolumn{3}{c|}{Loss weight $\alpha$ = 0.2} & \multicolumn{3}{c|}
        {Loss weight $\alpha$ = 0.4} & \multicolumn{3}{c}{Loss weight $\alpha$ = 0.6}\\
        \hline
         Acc.\% $\uparrow$& MSE$\downarrow$ & MAE$\downarrow$ & Acc.\% $\uparrow$& MSE$\downarrow$ & MAE$\downarrow$ & Acc.\% $\uparrow$& MSE$\downarrow$ & MAE$\downarrow$ \\
         \hline
         36.2& 1.24&0.83 & 38.6&1.26 & 0.82 & 32.4 & 1.70&0.99\\
 \bottomrule
    \end{tabular}
    }
    \end{center}

\end{table}

\begin{table}[t]
    \caption{The effect of two-stage fine-tuning on model performance. The model performance is measured by the best accuracy over three independent runs.}\label{pre-train}
\setlength{\tabcolsep}{3pt}
    \centering
    \begin{center}
          \resizebox{0.7\textwidth}{!}{
    \begin{tabular}{c|c|c|c}
    \toprule
    Model&w/o PICG & with PICG & Baseline MLLM\\
    \hline
Kang et al.~\cite{kang2024deep}&33.8\% &37.5\%   & 33.1\% \\
Sanford et al.~\cite{sanford2020deep}&30.7\% &36.5\% &29.7\% \\
Yu et al.~\cite{yu2023pi}&34.5\% &38.9\% &35.8\% \\
\bottomrule
    \end{tabular}
    }
    \end{center}
\end{table}

\section{Conclusion}
We introduce a novel method that integrates PICG into the PI-RADS scoring process via MLLM. Our approach involves a two-stage fine-tuning procedure to adapt MLLM to the MRI domain and comprehend PICG, supported by a domain adapter layer. Feature distillation integrates PICG into the scoring network. As a versatile plug-in for various networks, our model improves performance across three state-of-the-art scoring networks. Experiments with diverse demographics demonstrate that incorporating PICG enhances generalizability, likely due to improved network explainability, meriting further investigation.

\begin{credits}
\subsubsection{\ackname} This work was supported in part by the Hong Kong Research Grants Council (Projects No. T45-401/22-N and No. C4004-22Y), in part by Hong Kong Innovation and Technology Commission (Project No. ITS/200/22FP), in part by National Key R\&D Program of China Project 2022ZD0161100, and in part by DIREC project EXPLAIN-ME (9142-00001B). 

\subsubsection{\discintname}
The authors have no competing interests to declare that are relevant to the content of this article.
\end{credits}
\bibliographystyle{splncs04}
\bibliography{Paper-2830}

\begin{thebibliography}{10}
\providecommand{\url}[1]{\texttt{#1}}
\providecommand{\urlprefix}{URL }
\providecommand{\doi}[1]{https://doi.org/#1}

\bibitem{ahmed2017diagnostic}
Ahmed, H.U., Bosaily, A.E.S., Brown, L.C., Gabe, R., Kaplan, R., Parmar, M.K., Collaco-Moraes, Y., Ward, K., Hindley, R.G., Freeman, A., et~al.: Diagnostic accuracy of multi-parametric mri and trus biopsy in prostate cancer (promis): A paired validating confirmatory study. The Lancet  \textbf{389}(10071),  815--822 (2017)

\bibitem{alayrac2022flamingo}
Alayrac, J.B., Donahue, J., Luc, P., Miech, A., Barr, I., Hasson, Y., Lenc, K., Mensch, A., Millican, K., Reynolds, M., et~al.: Flamingo: A visual language model for few-shot learning. Advances in Neural Information Processing Systems  \textbf{35},  23716--23736 (2022)

\bibitem{carreira2019short}
Carreira, J., Noland, E., Hillier, C., Zisserman, A.: A short note on the kinetics-700 human action dataset. arXiv preprint arXiv:1907.06987  (2019)

\bibitem{carreira2017quo}
Carreira, J., Zisserman, A.: Quo vadis, action recognition? a new model and the kinetics dataset. In: Proceedings of the IEEE/CVF Conference on Computer Vision and Pattern Recognition. pp. 6299--6308 (2017)

\bibitem{bickle_2023}
Czarniecki~M, Bickle~I, W.Y.: Prostate imaging-reporting and data system (pi-rads): Radiology reference article. Website link: \url{https://radiopaedia.org/articles/prostate-imaging-reporting-and-data-system-pi-rads-1?lang=us} (2023)

\bibitem{dai2023instructblip}
Dai, W., Li, J., Li, D., Tiong, A., Zhao, J., Wang, W., Li, B., Fung, P., Hoi, S.: Instruct{BLIP}: Towards general-purpose vision-language models with instruction tuning. In: Thirty-seventh Conference on Neural Information Processing Systems (2023)

\bibitem{gao2023llama}
Gao, P., Han, J., Zhang, R., Lin, Z., Geng, S., Zhou, A., Zhang, W., Lu, P., He, C., Yue, X., et~al.: Llama-adapter v2: Parameter-efficient visual instruction model. arXiv preprint arXiv:2304.15010  (2023)

\bibitem{gravina2022machine}
Gravina, M., Spirito, L., Celentano, G., Capece, M., Creta, M., Califano, G., Coll{\`a}~Ruvolo, C., Morra, S., Imbriaco, M., Di~Bello, F., et~al.: Machine learning and clinical-radiological characteristics for the classification of prostate cancer in pi-rads 3 lesions. Diagnostics  \textbf{12}(7), ~1565 (2022)

\bibitem{gu2023deep}
Gu, W.j., Liu, Z., Yang, Y.j., Zhang, X.z., Chen, L.y., Wan, F.n., Liu, X.h., Chen, Z.z., Kong, Y.y., Dai, B.: A deep learning model, nafnet, predicts adverse pathology and recurrence in prostate cancer using mris. NPJ Precision Oncology  \textbf{7}(1), ~134 (2023)

\bibitem{kafkalias2022bias}
Kafkalias, A., Herodotou, S., Theodosiou, Z., Lanitis, A.: Bias in face image classification machine learning models: The impact of annotator’s gender and race. In: IFIP International Conference on Artificial Intelligence Applications and Innovations. pp. 89--100. Springer (2022)

\bibitem{kang2024deep}
Kang, Z., Xiao, E., Li, Z., Wang, L.: Deep learning based on resnet-18 for classification of prostate imaging-reporting and data system category 3 lesions. Academic Radiology  (2024)

\bibitem{koh2020concept}
Koh, P.W., Nguyen, T., Tang, Y.S., Mussmann, S., Pierson, E., Kim, B., Liang, P.: Concept bottleneck models. In: International conference on machine learning. pp. 5338--5348. PMLR (2020)

\bibitem{li2024llava}
Li, C., Wong, C., Zhang, S., Usuyama, N., Liu, H., Yang, J., Naumann, T., Poon, H., Gao, J.: Llava-med: Training a large language-and-vision assistant for biomedicine in one day. Advances in Neural Information Processing Systems  \textbf{36} (2024)

\bibitem{lin2022saw}
Lin, M., Feragen, A., Bashir, Z., Tolsgaard, M.G., Christensen, A.N.: I saw, i conceived, i concluded: Progressive concepts as bottlenecks. arXiv preprint arXiv:2211.10630  (2022)

\bibitem{liu2023functionconsistent}
Liu, D., Kan, M., Shan, S., CHEN, X.: Function-consistent feature distillation. In: The Eleventh International Conference on Learning Representations (2023)

\bibitem{monfort2019moments}
Monfort, M., Andonian, A., Zhou, B., Ramakrishnan, K., Bargal, S.A., Yan, T., Brown, L., Fan, Q., Gutfreund, D., Vondrick, C., et~al.: Moments in time dataset: One million videos for event understanding. IEEE transactions on pattern analysis and machine intelligence  \textbf{42}(2),  502--508 (2019)

\bibitem{natarajan2020prostate}
Natarajan, S., Priester, A., Margolis, D., Huang, J., Marks, L.: Prostate mri and ultrasound with pathology and coordinates of tracked biopsy (prostate-mri-us-biopsy). Cancer Imaging Arch  \textbf{10}, ~7937 (2020)

\bibitem{park2016prostate}
Park, S.Y., Jung, D.C., Oh, Y.T., Cho, N.H., Choi, Y.D., Rha, K.H., Hong, S.J., Han, K.: Prostate cancer: Pi-rads version 2 helps preoperatively predict clinically significant cancers. Radiology  \textbf{280}(1),  108--116 (2016)

\bibitem{purysko2021pi}
Purysko, A.S., Baroni, R.H., Giganti, F., Costa, D., Renard-Penna, R., Kim, C.K., Raman, S.S.: Pi-rads version 2.1: A critical review, from the ajr special series on radiology reporting and data systems. American Journal of Roentgenology  \textbf{216}(1),  20--32 (2021)

\bibitem{radford2021learning}
Radford, A., Kim, J.W., Hallacy, C., Ramesh, A., Goh, G., Agarwal, S., Sastry, G., Askell, A., Mishkin, P., Clark, J., et~al.: Learning transferable visual models from natural language supervision. In: International conference on machine learning. pp. 8748--8763. PMLR (2021)

\bibitem{sanford2020deep}
Sanford, T., Harmon, S.A., Turkbey, E.B., Kesani, D., Tuncer, S., Madariaga, M., Yang, C., Sackett, J., Mehralivand, S., Yan, P., et~al.: Deep-learning-based artificial intelligence for pi-rads classification to assist multiparametric prostate mri interpretation: A development study. Journal of Magnetic Resonance Imaging  \textbf{52}(5),  1499--1507 (2020)

\bibitem{schelb2019classification}
Schelb, P., Kohl, S., Radtke, J.P., Wiesenfarth, M., Kickingereder, P., Bickelhaupt, S., Kuder, T.A., Stenzinger, A., Hohenfellner, M., Schlemmer, H.P., et~al.: Classification of cancer at prostate mri: Deep learning versus clinical pi-rads assessment. Radiology  \textbf{293}(3),  607--617 (2019)

\bibitem{simonyan2014very}
Simonyan, K., Zisserman, A.: Very deep convolutional networks for large-scale image recognition. arXiv preprint arXiv:1409.1556  (2014)

\bibitem{wang2014principles}
Wang, Y.: Principles of magnetic resonance imaging: physics concepts, pulse sequences, \& biomedical applications. CreateSpace Independent Publishing (2014)

\bibitem{wu2023multimodal}
Wu, J., Gan, W., Chen, Z., Wan, S., Philip, S.Y.: Multimodal large language models: A survey. In: 2023 IEEE International Conference on Big Data (BigData). pp. 2247--2256 (2023)

\bibitem{yang2023language}
Yang, Y., Panagopoulou, A., Zhou, S., Jin, D., Callison-Burch, C., Yatskar, M.: Language in a bottle: Language model guided concept bottlenecks for interpretable image classification. In: Proceedings of the IEEE/CVF Conference on Computer Vision and Pattern Recognition. pp. 19187--19197 (2023)

\bibitem{yu2023pi}
Yu, R., Jiang, K.w., Bao, J., Hou, Y., Yi, Y., Wu, D., Song, Y., Hu, C.H., Yang, G., Zhang, Y.D.: Pi-radsai: Introducing a new human-in-the-loop ai model for prostate cancer diagnosis based on mri. British Journal of Cancer  \textbf{128}(6),  1019--1029 (2023)

\bibitem{zhang2023llama}
Zhang, R., Han, J., Zhou, A., Hu, X., Yan, S., Lu, P., Li, H., Gao, P., Qiao, Y.: Llama-adapter: Efficient fine-tuning of language models with zero-init attention. arXiv preprint arXiv:2303.16199  (2023)

\bibitem{zhang2023pmc}
Zhang, X., Wu, C., Zhao, Z., Lin, W., Zhang, Y., Wang, Y., Xie, W.: Pmc-vqa: Visual instruction tuning for medical visual question answering. arXiv preprint arXiv:2305.10415  (2023)

\end{thebibliography}

\end{document}